%% file: colm2024_conference.tex
\documentclass{article} %
\usepackage{colm2024_conference}

\usepackage{booktabs}
\usepackage{graphicx}
\usepackage{enumitem}
\usepackage{wrapfig}
\usepackage{algorithm}
\usepackage{algpseudocode}

\usepackage{microtype}
\usepackage{amsmath}
\usepackage{colortbl}
\usepackage[utf8]{inputenc}
\definecolor{lightgray}{rgb}{0.9,0.9,0.9}
\usepackage{caption}
\usepackage{subcaption}
\usepackage{xcolor}
\usepackage{setspace}
\usepackage{url}
\usepackage{multirow}
\usepackage{colortbl}
\usepackage{tabularx}
\usepackage{blindtext}
\usepackage{pgfplots}
\pgfplotsset{compat=1.18} 
\usepackage{tikz}
\usetikzlibrary{er,positioning,bayesnet}
\usepackage{makecell}
\usepackage{tipa}
\usepackage{siunitx}
\usepackage{nicefrac}
\usepackage{tocloft}
\usepackage{listings}
\usepackage[raster,skins]{tcolorbox} %
\usepackage{xltabular}
\usepackage{adjustbox}
\usepackage{xurl}
\usepackage{cleveref}

\usepackage{float}
\usepackage{wrapfig}
\usepackage{subcaption}

\input{math_commands.tex}

\title{\textit{FastMTP:} Accelerating LLM Inference with \\Enhanced Multi-Token Prediction}

\author{
  \bf Yuxuan Cai, Xiaozhuan Liang, Xinghua Wang, Jin Ma, Haijin Liang, Jinwen Luo, Xinyu Zuo, Lisheng Duan, Yuyang Yin, Xi Chen\\
  \vspace{0.5em}
  Tencent
}
\begin{document}

\maketitle

\begin{abstract}
As large language models (LLMs) become increasingly powerful, the sequential nature of autoregressive generation creates a fundamental throughput bottleneck that limits the practical deployment. While Multi-Token Prediction (MTP) has demonstrated remarkable benefits for model training efficiency and performance, its inherent potential for inference acceleration remains largely unexplored. This paper introduces FastMTP, a simple yet effective method that improves multi-step draft quality by aligning MTP training with its inference pattern, significantly enhancing speculative decoding performance. Our approach fine-tunes a single MTP head with position-shared weights on self-distilled data, enabling it to capture dependencies among consecutive future tokens and maintain high acceptance rates across multiple recursive draft steps. By integrating language-aware dynamic vocabulary compression into the MTP head, we further reduce computational overhead in the drafting process. Experimental results across seven diverse benchmarks demonstrate that FastMTP achieves an average of \textbf{2.03$\times$ speedup} compared to standard next token prediction with lossless output quality, outperforming vanilla MTP by 82\%. FastMTP requires only lightweight training and seamlessly integrates with existing inference frameworks, offering a practical and rapidly deployable solution for accelerating LLM inference.
\end{abstract}

\vspace{-15pt}

% \newpage

\section{Introduction}
% ~\citep{deng2025emerging, liao2025mogao}.

Large Language Models (LLMs) have demonstrated remarkable capabilities~\citep{deepseek-ai2025deepseekr1,yang2025qwen3,glm-4.5team2025glm45,kimiteam2025kimi} across diverse applications including autonomous agents~\citep{wangSurveyLargeLanguage2024,guoLargeLanguageModel2024}, code generation~\citep{liuLargeLanguageModelBased2024,jiangSurveyLargeLanguage2024}, and complex reasoning tasks~\citep{chenReasoningEraSurvey2025, ahnLargeLanguageModels2024}. However, their practical deployment faces a fundamental efficiency bottleneck: the autoregressive nature of token generation. Current LLMs generate text sequentially, producing only one token per forward pass, which means the overall generation time scales linearly with sequence length~\citep{santilli2023accelerating}. This becomes particularly problematic for scenarios requiring extensive generation, such as the state-of-the-art large reasoning models~\citep{openai2024openai,deepseek-ai2025deepseekr1}, which have achieved breakthrough progress in solving complex and logic-intensive tasks by generating extended human-like Chain-of-Thoughts (CoTs)~\citep{wei2022chainofthought,sprague2025cot} before reaching a final answer. While these models demonstrate strong reasoning capability, they often produce excessively long reasoning chains, even for simple samples, inevitably introducing substantial computational overhead~\citep{feng2025efficient}. To this end, it highlights an urgent need for effective acceleration.

Recent research has explored multiple strategies to accelerate LLM inference, including efficient attention mechanisms~\citep{katharopoulos2020transformers,child2019generating,yang2024gated,lu2025moba,dao2023flashattention2} and model compression~\citep{lin2024awq,xiao2024smoothquant,gu2024minillm,hsieh2023distilling} to reduce computational overhead. Some works have also focused on reducing CoT redundancy, including reinforcement learning with length penalties~\citep{luo2025o1pruner,aggarwal2025l1} and supervised fine-tuning on variable-length CoT data~\citep{ma2025cotvalve}. Among these approaches, speculative decoding~\citep{leviathan2023fast,chen2023accelerating,stern2018blockwise} has emerged as a promising technique that enhances decoding efficiency without compromising the fidelity of outputs. The core idea of this approach involves employing a smaller model, termed a draft model, to predict several subsequent tokens that are then verified by the target LLM in parallel, achieving multi-token generation per forward pass~\citep{zhou2024survey}.

% , initially proposed by Meta

Multi-Token Prediction (MTP)~\citep{gloeckle2024better} modules and their auxiliary training, originally designed to improve training, offer a natural opportunity for inference acceleration through speculative decoding. MTP extends the traditional next-token prediction paradigm by training language models to predict multiple future tokens simultaneously. This approach encourages models to plan ahead and leverages richer supervision signals. Building on this foundation, DeepSeek-V3~\citep{deepseek-ai2024deepseekv3} refined the MTP architecture with a sequential implementation using cascaded MTP modules, preserving the complete causal chain to maintain the autoregressive nature.
% Such designs are now widely adopted across state-of-the-art models~\citep{qwenteam2025qwen3next80ba3binstruct,glm-4.5team2025glm45,meituanlongcatteam2025longcatflash,xiaomi2025mimo}.

% MTP extends the traditional next-token prediction paradigm by training language models to predict multiple future tokens simultaneously. This approach encourages models to plan ahead and leverages richer supervision signals. Building on this foundation, DeepSeek-V3 refined the MTP architecture with a sequential implementation that preserves the complete causal chain to maintain the autoregressive nature~\citep{deepseek-ai2024deepseekv3}. Their approach utilizes the main model's hidden states combined with shifted token embeddings to predict additional tokens through cascaded MTP modules, where each module sequentially predicts one future position based on the previous module's output.

% In principle, these lightweight MTP modules present a natural opportunity for inference acceleration through speculative decoding~\citep{stern2018blockwise,leviathan2023fast}: they can serve as draft models to rapidly generate multiple candidate tokens for parallel verification by the main model, promising significant reductions in generation latency. 

% The adoption of MTP modules as an auxiliary training objective is emerging as a promising trend in state-of-the-art language models~\citep{qwenteam2025qwen3next80ba3binstruct,glm-4.5team2025glm45,meituanlongcatteam2025longcatflash,xiaomi2025mimo}. However, 

Despite growing adoption for training improvement~\citep{qwenteam2025qwen3next80ba3binstruct,glm-4.5team2025glm45,meituanlongcatteam2025longcatflash,xiaomi2025mimo}, the potential of MTP for inference acceleration remains largely unexploited. Current implementations either discard the MTP modules entirely during inference~\citep{deepseek-ai2024deepseekv3}, reverting to standard next-token prediction, or keep only the first MTP module for multi-token prediction. This underutilization in inference may stem from two fundamental challenges in existing MTP implementations. First, the sequential MTP architecture requires cascaded forward passes through multiple MTP modules, with each module maintaining separate weights and key-value caches, resulting in substantial memory overhead. Furthermore, while many inference frameworks now support speculative decoding with a single draft model, MTP's design necessitates loading and orchestrating multiple draft models—one MTP module for each prediction step—requiring complex scheduling that severely impacts computational efficiency. This may explain why models trained with multiple MTP layers often open-source only a single module for inference. Second, attempts to circumvent this complexity by recursively reusing a single MTP module yield poor acceptance rates beyond the first additional token, as this module was not explicitly trained for recursive multi-step prediction patterns, making it ineffective for extended draft generation.
% , failing to leverage the full multi-step generation potential.

In this paper, we present FastMTP, an enhanced multi-token prediction framework that makes the MTP module more effective, efficient, and deployable during inference. Our approach fine-tunes a single MTP head with shared weights across all prediction steps, teaching it to perform multi-token generation while maintaining causality. This enables the model to capture dependencies among consecutive future tokens, resulting in higher acceptance rates beyond the initial draft position, and ensures compatibility with EAGLE-style speculative decoding~\citep{li2024eagle} for seamless integration with existing inference frameworks such as SGLang~\citep{zheng2024sglang}. Inspired by FR-Spec~\citep{zhao2025frspec}, we integrate language-aware dynamic vocabulary compression that further reduces the computational cost of draft generation with negligible impact on acceptance rates across diverse tasks and languages.
%  through computationally inexpensive post-training， and adapting it with language-specific frequency analysis

Our key contributions are as follows:
\begin{itemize}
    \item We adapt a single MTP head to perform effective recursive multi-step draft generation through fine-tuning on self-distilled data, dramatically improving average acceptance rates from 70\% to 81\% for the first draft token, from 11\% to 56\% for the second draft token and from 2\% to 36\% for the third token compared to the vanilla MTP reusing strategy, unlocking its full potential for inference acceleration.

    \item We adopt language-aware dynamic vocabulary compression for the MTP head, which reduces computational overhead during drafting according to the input context. It further increases average output throughput by approximately 16\% when drafting three additional tokens.
    
    \item We conduct extensive experiments on 7 benchmarks, demonstrating an average 2.03$\times$ speedup on 7B models with lossless generation quality, significantly outperforming the vanilla MTP reusing strategy (1.21$\times$).
    
    % \item We open-source FastMTP checkpoints based on MiMo-7B-RL~\citep{xiaomi2025mimo}, training and inference code, along with pre-computed vocabulary statistics, enabling full reproducibility and rapid deployment.
    
\end{itemize}

% \section{Preliminary}

% \subsection{Multi-token Prediction} \label{sec:mtp_pre}
% Next token prediction paradigm has been prevailing for autoregressive models in the era of LLMs
% Multi-token prediction architecture and implementation

% \subsection{Speculative Decoding}
% Drafting and verification

% Speculative decoding is founded on two key insights into the inference dynamics of LLMs. First, generating tokens with a smaller draft model reduces computational cost by offloading initial token prediction from the larger target model. Second, LLM inference is primarily constrained by memory bandwidth, with latency bottlenecks arising from parameter memory access rather than arithmetic operations
% Despite its potential, speculative decoding raises several challenges that require deeper exploration. Central among these is the need to design an effective drafting mechanism that optimally balances speculative accuracy with drafting efficiency
% The effectiveness of this approach depends on two critical factors: the accuracy of the draft model, measured by the average number of accepted tokens per decoding step, and the drafting latency itself
% Achieving a trade-off between high speculative accuracy and low latency remains a significant challenge, as both elements are essential for maximizing the overall speedup.

\section{Methodology}

\begin{figure}[htbp]
    \centering
    \includegraphics[width=1.0\linewidth]{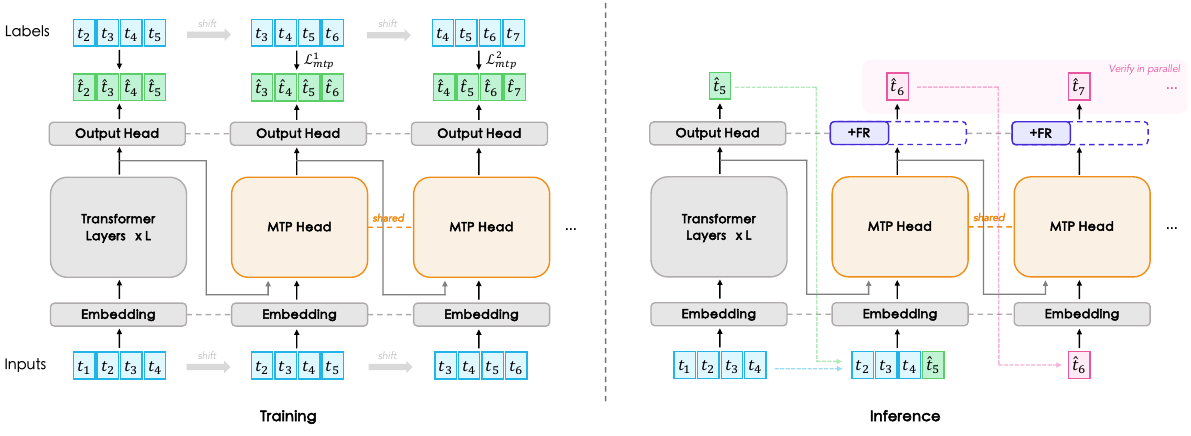}
    \caption{Illustration of FastMTP training and inference strategy. \textbf{Training phase (left):} Grey blocks represent frozen main model modules, orange blocks denote trainable MTP heads with shared weights, and blue blocks show input and label sequences with sequential position shifts. \textbf{Inference phase (right):} The main model predicts the next token (green), which feeds into the MTP head for recursively generating draft tokens (pink) with parallel verification. Purple blocks indicate frequency-ranked (FR) language-aware dynamic vocabulary compression that accelerates draft generation.}
    \label{fig:overview}
\end{figure}

In this section, we provide a detailed description of the implementation of FastMTP, which enhances MTP specifically for more effective and efficient speculative decoding during LLM inference. Figure \ref{fig:overview} illustrates its training and inference pipelines.

%  for autoregressive training,top,bottom

\subsection{Training of the shared MTP head} \label{sec:train_recipe}

% , as detailed in Section \ref{sec:mtp_pre}
\paragraph{Architecture Design.} FastMTP adopts the same MTP architecture as DeepSeek-V3~\citep{deepseek-ai2024deepseekv3}. The key distinction lies in our use of a single MTP head with shared weights across all prediction steps, departing from the conventional approach that employs multiple independent modules for each prediction depth. This shared-weight design not only reduces memory usage, but more importantly, forces the model to capture dependencies among consecutive future tokens for causal multi-token prediction.

\paragraph{Training Mechanism.} 

Let $\mathcal{F}$ denote the transformer layers of the main model and $\mathcal{M}$ denote the MTP head. Consider processing tokens at position $i$ in a sequence. The input tokens $t_{1:i}$ first pass through the embedding layer and transformer layers $\mathcal{F}$ to produce hidden states $h_{1:i}$. For predicting $K$ additional future tokens, the MTP head $\mathcal{M}$ operates recursively:

\begin{itemize}

    \item At step $k=1$: $\mathcal{M}$ takes the hidden state $h_i$ from $\mathcal{F}$ along with the embedding of the shifted token $t_{i+1}$ (shifted by 1 position) to predict token $\hat{t}^1_{i+2}$.
    
    \item At step $k>1$: $\mathcal{M}$ processes its own output hidden state $h^{k-1}_{i+k-1}$ from step $k-1$, combined with the embedding of the shifted token $t_{i+k}$ (shifted by $k$ positions), to predict token $\hat{t}^k_{i+k+1}$.
    
\end{itemize}

During training, this process is applied to all valid positions in the sequence. For a training sequence of length $T$, position $i$ ranges from $1$ to $T-K$ to ensure all indices remain valid after shifting during the $K$ prediction steps.

\paragraph{Training Objective.} We optimize the MTP head using a weighted cross-entropy loss with exponential decay for distant token predictions:

\begin{equation}
    \mathcal{L}_{mtp} = \sum_{k=1}^{K} \alpha_k \cdot \mathcal{L}_{mtp}^k = \sum_{k=1}^{K} \alpha_k \cdot \text{CE}\,(\hat{t}^k_{1+k\,:\,T+1}, \,t_{1+k\,:\,T+1})
\end{equation}

where $\mathcal{L}_{mtp}^k$ represents the loss for the $k$-th prediction step, and $\text{CE}(\cdot, \cdot)$ denotes the cross-entropy loss between predicted and ground truth tokens. The position-dependent weights $\alpha_k$ follow an exponential decay:

\begin{equation}
    \alpha_k = \frac{\beta^{k-1}}{\sum_{j=1}^{K} \beta^{j-1}}
\end{equation}

where $\beta$ is the decay factor. This weighting strategy considers that the prediction uncertainty accumulates with each additional step - distant tokens depend on more intermediate decisions, making them progressively harder to predict. The exponential decay ensures the model prioritizes near-term predictions while still developing generation capabilities for several sequential future tokens.

Only the MTP head $\mathcal{M}$ is fine-tuned while all main model components remain frozen, updating less than 3\% of parameters, and ensuring both computational efficiency and preservation of the base model's capabilities.

\subsection{Language-aware vocabulary space compression in MTP drafting}

To accelerate the drafting phase, we combine frequency-based vocabulary compression with MTP, reducing the computational overhead of the MTP output head. Following the analysis in FR-Spec~\citep{zhao2025frspec}, a small subset of tokens accounts for the vast majority of occurrences while the remaining tokens exhibit extremely sparse frequencies—a consistent long-tail pattern. This observation motivates us to restrict the MTP head's output space to high-frequency tokens during draft generation. 

Since the original vocabulary compression methods primarily analyzed token distributions on English corpora, their compressed vocabularies poorly represent Chinese tokens, which we find severely limits performance on Chinese downstream tasks. To overcome this, we compute language-specific frequency statistics and dynamically adjust high-frequency vocabularies based on the generation context, ensuring adequate high-frequency tokens for different languages are represented in the compressed vocabulary space. This language-aware compression maintains high draft acceptance rates for both English and Chinese generation while preserving computational efficiency.

Let $\mathcal{V}$ denote the full vocabulary of the language model. For each language $l$, we define $\mathcal{V}_{high}^{(l)} \subset \mathcal{V}$ as the subset of high-frequency tokens specific to that language, identified through corpus-level statistics. During draft generation, the MTP head dynamically selects the appropriate vocabulary based on the current context. In other words, if the main model's output head is $\mathbf{W}\in\mathbb{R}^{|\mathcal{V}|\times d}$, we extract the language-specific submatrix:

\begin{equation}
    \tilde{\mathbf{W}}^{(l)}[i,:]=\mathbf{W}[\mathcal{V}_{high}^{(l)}[i],\,:],\quad i=1, ...,|\mathcal{V}_{high}^{(l)}|
\end{equation}

where the language $l$ is determined based on the input context, enabling focused compression that targets language-specific high-frequency patterns. With the extracted $\tilde{\mathbf{W}}^{(l)}$, the MTP head computes output logits only for tokens in $\mathcal{V}_{high}^{(l)}$ rather than the full vocabulary, reducing computational cost while maintaining high acceptance rates across different languages. Note that the vocabulary compression technique is restricted to the drafting phase, the verification phase retains the full vocabulary space, guaranteeing lossless generation quality.

\subsection{EAGLE-Style MTP inference}

The inference process differs from training by using autoregressively generated tokens from previous steps rather than teacher forcing. Given an input token sequence $t_{1:i}$, inference proceeds through two phases: draft generation and parallel verification.

\paragraph{Drafting phase.} In the draft generation phase, the input first passes through the main model to produce the next token $\hat{t}_{i+1}$, which serves as the first verified token. The MTP head $\mathcal{M}$ then autoregressively generates $K$ draft tokens following the EAGLE methodology~\citep{li2024eagle}:

\begin{itemize}
    \item At the initial draft step $k=1$: 
    $\mathcal{M}$ takes the last hidden states from the transformer layers $\mathcal{F}$ along with the embedding of the concatenated sequence $[t_{2:i}; \hat{t}_{i+1}]$, where the newly predicted token is appended to the input sequence shifted one step ahead. This produces the first draft token $\hat{t}_{i+2}$.

    \item At the subsequent draft step $k>1$:
    $\mathcal{M}$ operates autoregressively, processing its own output hidden state $\hat{h}_{i+k}$ from step $k-1$ combined with the embedding of the previously drafted token $\hat{t}_{i+k}$.
\end{itemize}

After $K$ recursive steps, we obtain the complete draft sequence $\hat{t}_{i+2:i+K+1}$.

\paragraph{Verification phase.} The main model then processes all draft tokens in parallel, computing logits for positions $i+2$ through $i+K+1$ simultaneously. We adopt standard speculative decoding acceptance criteria to determine the number of accepted draft tokens: tokens are accepted sequentially until the first position where the draft token differs from what the main model would have sampled. Through parallel verification without approximate strategies, this ensures that the output distribution remains identical to the original model, which has been theoretically proven~\citep{leviathan2023fast,chen2023accelerating}, while achieving speedup through the acceptance of multiple tokens per forward pass.

\subsection{Training data}

We employ a self-distillation approach for training data generation, where the main model itself generates all training responses. Inspired by prior works~\citep{yang2024selfdistillation,cai2024medusa}, this strategy ensures natural alignment between the MTP head and the main model's distribution, leading to higher acceptance rates and more effective draft generation during speculative decoding.

Specifically, we collect diverse prompts from instruction-tuning datasets spanning multiple domains and languages. For each original sample $(x_n, y_n)$ where $x_n$ is the prompt and $y_n$ is the corresponding response from the source dataset, we generate a new response $\tilde{y}_n$ using the main model. The resulting self-distilled dataset $\{(x_n, \tilde{y}_n)\}$ captures the main model's semantic characteristics, generation patterns, and preferences. Training the MTP head on this self-distilled data ensures it learns to produce draft tokens consistent with the main model's behavior, rather than attempting to mimic responses from external sources that may have different distributions.

Our final training corpus comprises 389.4K samples in English and Chinese, spanning general knowledge, mathematical reasoning, and coding tasks. The detailed domain distribution, along with dataset curation and filtering procedures, is provided in Appendix \ref{app:train_data}.

\section{Experiments}

\subsection{Experimental Setup}

\paragraph{Model Configuration.} 
FastMTP is implemented using the pre-trained MiMo-7B-RL checkpoint~\citep{xiaomi2025mimo}, a dense 7B parameter model with 36 decoder layers, a single-layer MTP module, and a vocabulary size of 152K tokens (based on the Qwen2tokenizer~\citep{yang2024qwen2}). We adopt a computationally efficient strategy that freezes the main model (including the transformer layers, embedding layer, and output head), fine-tuning only the MTP head's 210.8M parameters, which accounts for less than 3\% of the backbone model's 7,833.4M total parameters.

\paragraph{Training Details.} 
The MTP head was trained for 3 epochs on 389.4K self-distilled samples (see Appendix \ref{app:train_data} for dataset details). Training employed cosine learning rate scheduling with a peak learning rate of 5e-5 and a warmup ratio of 0.05. The AdamW optimizer parameters were set to $(\beta_1, \beta_2) = (0.9,\, 0.95)$, with the global batch size set to 64. For the MTP-specific hyperparameters, we used a loss weight decay factor $\beta = 0.6$ and prediction depth $K = 3$. Training was conducted using the ms-swift framework~\citep{zhao2025swift}. The entire training process was completed in less than 1 day on a single H20 server, demonstrating low training cost.
% typo! \beta is 0.6

\paragraph{Evaluation Method.} 
We evaluate FastMTP on seven tasks adapted from Spec-Bench~\citep{xia2024unlocking}: MT-Bench~\citep{zheng2023judging} for multi-turn conversation, LiveCodeBench-v6~\citep{jain2024livecodebench} for coding, MATH-500~\citep{lightman2023lets} for mathematical reasoning, Natural Questions~\citep{kwiatkowski2019natural} for both RAG and question answering, CNN/Daily Mail~\citep{nallapati2016abstractive} for summarization, and C-Eval~\citep{huang2023ceval} for Chinese knowledge assessment. Details of these evaluation benchmarks are provided in Appendix \ref{app:bench}. All evaluations employ strict speculative decoding acceptance criteria, ensuring no degradation in generation quality. All experiments were conducted using the SGLang inference framework~\citep{zheng2024sglang} with single-batch inference, greedy decoding (temperature 0), and a maximum generation length of 1024 tokens across all tasks.

\paragraph{Metrics.} We use four widely used metrics for evaluation: (1) Average acceptance length $\tau$: mean accepted tokens per forward pass of the main model. (2) Acceptance Rate: the percentage of draft tokens accepted during verification. (3) Average output throughput: mean output tokens per second (token/s). (4) Speedup ratio: relative speedup compared to baseline (vanilla autoregressive decoding).

\paragraph{Hardware Settings.} 
We conducted our primary experiments on an NVIDIA A10 GPU (24GB), a widely-deployed accelerator in production environments. 

% Further analyses in Section \ref{sec:draft_len} were performed on a more powerful NVIDIA A100 GPU (40GB).

Note that speculative decoding theoretically preserves the main model's output distribution~\citep{leviathan2023fast,chen2023accelerating}. Since we employ strict acceptance criteria without any relaxation, the outputs are identical to vanilla autoregressive decoding, making separate generation quality evaluation unnecessary.

\subsection{Main Results}

We evaluate three primary configurations: (1) vanilla autoregressive decoding \textit{(baseline)}, (2) the original MTP checkpoint without fine-tuning \textit{(vanilla MTP)}, and (3) the proposed FastMTP. To analyze the contribution of each component, we further evaluate FastMTP variants: 
\begin{itemize}
    \item Fixed-data FT: Fine-tunes using both prompts and responses from original data sources.

    \item Self-data FT: Employs self-distilled responses generated by the main model.

    \item Self-data FT + FR: Additionally incorporates language-aware vocabulary compression.
\end{itemize}
We assess performance across different draft depths $K \in \{0, 1, 2, 3\}$, where $K=0$ represents vanilla autoregressive decoding without drafting, and $K=3$ indicates three recursive MTP forward passes generating three additional draft tokens.

% This represents an 82\% improvement over vanilla MTP (1.21$\times$), a 36\% gain over fixed-data fine-tuning (1.67$\times$), and a 22\% increase over self-distilled data fine-tuning without vocabulary compression (1.81$\times$). These incremental improvements validate the effectiveness of each component: fine-tuning the shared-weight MTP head enables more accurate and faster multi-token drafting during inference, self-distillation achieves better distribution alignment between draft and main model for higher acceptance length, and vocabulary compression reduces computational overhead to increase output throughput with minimal impact on acceptance rates.

\begin{table}[ht]
\centering
\caption{Average Accepted Length and Decoding Speed (token/s) for MiMo-RL-7B Using Various Methods to Predict K Draft Tokens. Tasks include multi-turn conversation (MT.), coding (Code), mathematical reasoning (Math), retrieval-augmented generation (RAG), question answering (QA), summarization (Summ.), and Chinese knowledge (ZH.)}
\small
\begin{adjustbox}{width=\textwidth}
\label{tab:all}
\begin{tabular}{clcccccccccccccccc}
\toprule
                                  & \multicolumn{1}{c}{}                          & \multicolumn{2}{c}{\textbf{MT.}}                                               & \multicolumn{2}{c}{\textbf{Code}}                                              & \multicolumn{2}{c}{\textbf{Math}}                                              & \multicolumn{2}{c}{\textbf{RAG}}                                               & \multicolumn{2}{c}{\textbf{QA}}                                                & \multicolumn{2}{c}{\textbf{Summ.}}                                             & \multicolumn{2}{c}{\textbf{ZH.}}                                               & \multicolumn{2}{c}{\textbf{Mean}}                                                      \\ \cmidrule{3-18} 
\multirow{-2}{*}{} & \multicolumn{1}{c}{\multirow{-2}{*}{\textbf{Method}}}  & $\tau$                            & token/s                                & $\tau$                            & token/s                                & $\tau$                            & token/s                                & $\tau$                            & token/s                                & $\tau$                            & token/s                                & $\tau$                            & token/s                                & $\tau$                            & token/s                                & $\tau$                            & token/s                              \\ \midrule
K=0                               & Baseline                                      & 1.00                         & 31.28                                  & 1.00                         & 31.49                                  & 1.00                         & 31.90                                  & 1.00                         & 31.27                                  & 1.00                         & 31.92                                  & 1.00                         & 31.42                                  & 1.00                         & 31.63                                  & 1.00                         & 31.55 (1.00x)                                  \\ \midrule
                                  & Vanilla MTP                                   & 1.72                         & 41.86                                  & 1.73                         & 42.71                                  & 1.83                         & 44.41                                  & 1.65                         & 40.22                                  & 1.67                         & 41.06                                  & 1.64                         & 39.78                                  & 1.67                         & 41.66                                  & 1.70                         & 41.67 (1.32x)                                  \\
                                  & Fixed-data FT                             & 1.75                         & 43.46                                  & 1.80                         & 44.67                                  & 1.86                         & 45.06                                  & 1.78                         & 43.17                                  & 1.71                         & 41.85                                  & 1.72                         & 42.40                                  & 1.74                         & 42.70                                  & 1.76                         & 43.33 (1.37x)                                  \\
                                  & Self-data FT                             & 1.80                         & 44.51                                  & 1.84                         & 45.39                                  & 1.90                         & 46.43                                  & 1.82                         & 43.99                                  & 1.74                         & 42.67                                  & 1.77                         & 42.97                                  & 1.76                         & 44.30                                  & 1.81                         & 44.32 (1.40x)                                  \\
\multirow{-4}{*}{K=1}             & Self-data FT + FR                         & 1.78                         & 45.79                                  & 1.82                         & 47.59                                  & 1.88                         & 47.58                                  & 1.80                         & 45.66                                  & 1.74                         & 44.60                                  & 1.74                         & 45.45                                  & 1.76                         & 46.38                                  & 1.79                         & 46.15 (1.46x)                                  \\ \midrule
                                  & Vanilla MTP                                   & 1.85                         & 42.03                                  & 1.82                         & 41.70                                  & 2.00                         & 45.52                                  & 1.71                         & 38.35                                  & 1.82                         & 41.26                                  & 1.73                         & 39.00                                  & 1.78                         & 41.27                                  & 1.81                         & 41.30 (1.31x)                                  \\
                                  & Fixed-data FT                             & 2.20                         & 49.83                                  & 2.33                         & 53.34                                  & 2.52                         & 56.51                                  & 2.31                         & 51.65                                  & 2.10                         & 47.47                                  & 2.14                         & 47.29                                  & 2.19                         & 49.85                                  & 2.26                         & 50.85 (1.61x)                                  \\
                                  & Self-data FT                              & 2.35                         & 53.90                                  & 2.45                         & 55.67                                  & 2.62                         & 59.68                                  & 2.40                         & 53.96                                  & 2.23                         & 50.71                                  & 2.26                         & 50.93                                  & 2.27                         & 52.66                                  & 2.37                         & 53.93 (1.71x)                                  \\
\multirow{-4}{*}{K=2}             & Self-data FT + FR                         & 2.30                         & 58.12                                  & 2.40                         & 60.44                                  & 2.59                         & 63.71                                  & 2.37                         & 58.63                                  & 2.21                         & 55.47                                  & 2.19                         & 55.02                                  & 2.24                         & 58.20                                  & 2.33                         & 58.51 (1.85x)                                  \\ \midrule
                                  & Vanilla MTP                                   & 1.86                         & 38.97                                  & 1.83                         & 38.17                                  & 2.02                         & 42.06                                  & 1.72                         & 35.27                                  & 1.83                         & 38.10                                  & 1.75                         & 36.17                                  & 1.78                         & 37.58                                  & 1.83                         & 38.04 (1.21x)                                  \\
                                  & Fixed-data FT                             & 2.48                         & 51.53                                  & 2.64                         & 55.40                                  & 2.93                         & 59.72                                  & 2.62                         & 53.97                                  & 2.30                         & 48.08                                  & 2.35                         & 48.38                                  & 2.45                         & 51.69                                  & 2.54                         & 52.68 (1.67x)                                  \\
                                  & Self-data FT                              & \textbf{2.69}                & 56.33                                  & \textbf{2.85}                & 59.49                                  & \textbf{3.16}                & 65.94                                  & \textbf{2.80}                & 57.55                                  & \textbf{2.49}                & 52.31                                  & \textbf{2.55}                & 53.08                                  & \textbf{2.55}                & 54.36                                  & \textbf{2.73}                & 57.01 (1.81x)                                  \\
\multirow{-4}{*}{K=3}             & \cellcolor[HTML]{E5F6FF}Self-data FT + FR & \cellcolor[HTML]{E5F6FF}2.62 & \cellcolor[HTML]{E5F6FF}\textbf{63.42} & \cellcolor[HTML]{E5F6FF}2.75 & \cellcolor[HTML]{E5F6FF}\textbf{66.24} & \cellcolor[HTML]{E5F6FF}3.07 & \cellcolor[HTML]{E5F6FF}\textbf{73.66} & \cellcolor[HTML]{E5F6FF}2.75 & \cellcolor[HTML]{E5F6FF}\textbf{64.55} & \cellcolor[HTML]{E5F6FF}2.46 & \cellcolor[HTML]{E5F6FF}\textbf{58.88} & \cellcolor[HTML]{E5F6FF}2.47 & \cellcolor[HTML]{E5F6FF}\textbf{59.15} & \cellcolor[HTML]{E5F6FF}2.53 & \cellcolor[HTML]{E5F6FF}\textbf{62.93} & \cellcolor[HTML]{E5F6FF}2.66 & \cellcolor[HTML]{E5F6FF}\textbf{64.12 (2.03x)} \\ \bottomrule
\end{tabular}
\end{adjustbox}
\end{table}

% \begin{figure}[htbp]
%     \centering
%     \includegraphics[width=0.6\linewidth]{./figures/radar_chart.pdf}
%     \caption{Speedup comparison of different methods across subtasks, evaluated on a single A10 GPU.}
%     \label{fig:radar}
% \end{figure}

\begin{wrapfigure}{r}{0.5\textwidth}  % r=右侧, l=左侧, 0.4=占40%宽度
    \centering
    \includegraphics[width=0.48\textwidth]{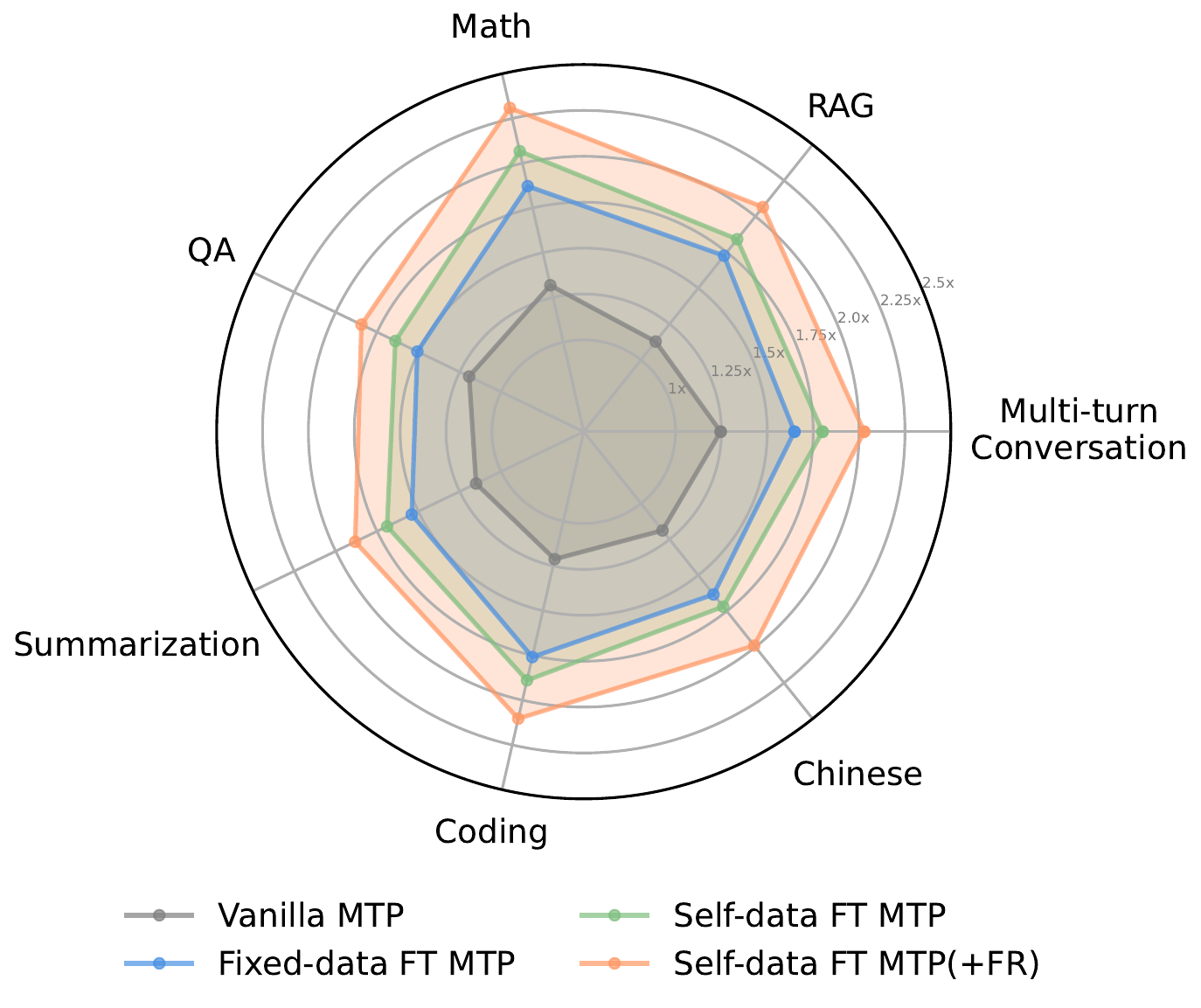}
    \caption{Speedup comparison of different methods across subtasks, evaluated on a single A10 GPU.}
    \label{fig:radar}
\end{wrapfigure}

Table \ref{tab:all} and Figure \ref{fig:radar} present comprehensive acceleration performance evaluations of FastMTP across seven diverse tasks. FastMTP with self-distillation and vocabulary compression \textit{(Self-data FT + FR)} achieves superior performance across all benchmarks, delivering an average 2.03$\times$ speedup over vanilla autoregressive decoding at $K=3$. 
The performance gains vary across task domains, reflecting their distinct generation characteristics. Mathematical reasoning exhibits the highest speedup and average acceptance length ($\tau=3.16$ at $K=3$ before vocabulary compression), demonstrating the MTP head's effectiveness in capturing structured reasoning patterns. Coding tasks deliver the second-best performance, benefiting from prevalent fixed templates and repetitive programming constructs. General NLP tasks such as question answering show comparatively smaller improvements (1.84$\times$--2.07$\times$), potentially due to their higher linguistic diversity and less predictable token dependencies. This task-specific variation, visualized in Figure \ref{fig:radar}, confirms that FastMTP maintains consistent acceleration benefits across diverse generation scenarios.

% \begin{figure}
%     \centering
%     \includegraphics[width=0.9\linewidth]{./figures/acc_rate.pdf}
%     \caption{Acceptance rate improvement of self-distilled dataset fine-tuned MTP compared with vanilla MTP at different draft steps.}
%     \label{fig:acc_rate}
% \end{figure}

\subsection{Further Analyses}

Our further analyses explore two critical aspects of FastMTP's design and performance:

(1) Optimal draft length: What is the optimal number of MTP draft steps for achieving maximum speedup? In other words, what is the effective prediction distance that a single MTP head can learn while maintaining high acceptance rates?

(2) Vocabulary compression trade-offs: What degree of vocabulary compression achieves the optimal balance between computational efficiency and acceptance rate? How does this trade-off vary across different domains and languages?

\subsubsection{Optimal draft length} \label{sec:draft_len}

To determine the optimal number of draft steps, we trained an MTP head capable of predicting up to 7 additional tokens, applying the same exponentially decaying loss weighting strategy described in Section \ref{sec:train_recipe}. We then evaluated the decoding speed and acceptance length for different values of drafting step $K$.
% on a single A100 GPU.

\begin{figure}[t]
    \centering
    \includegraphics[width=0.8\linewidth]{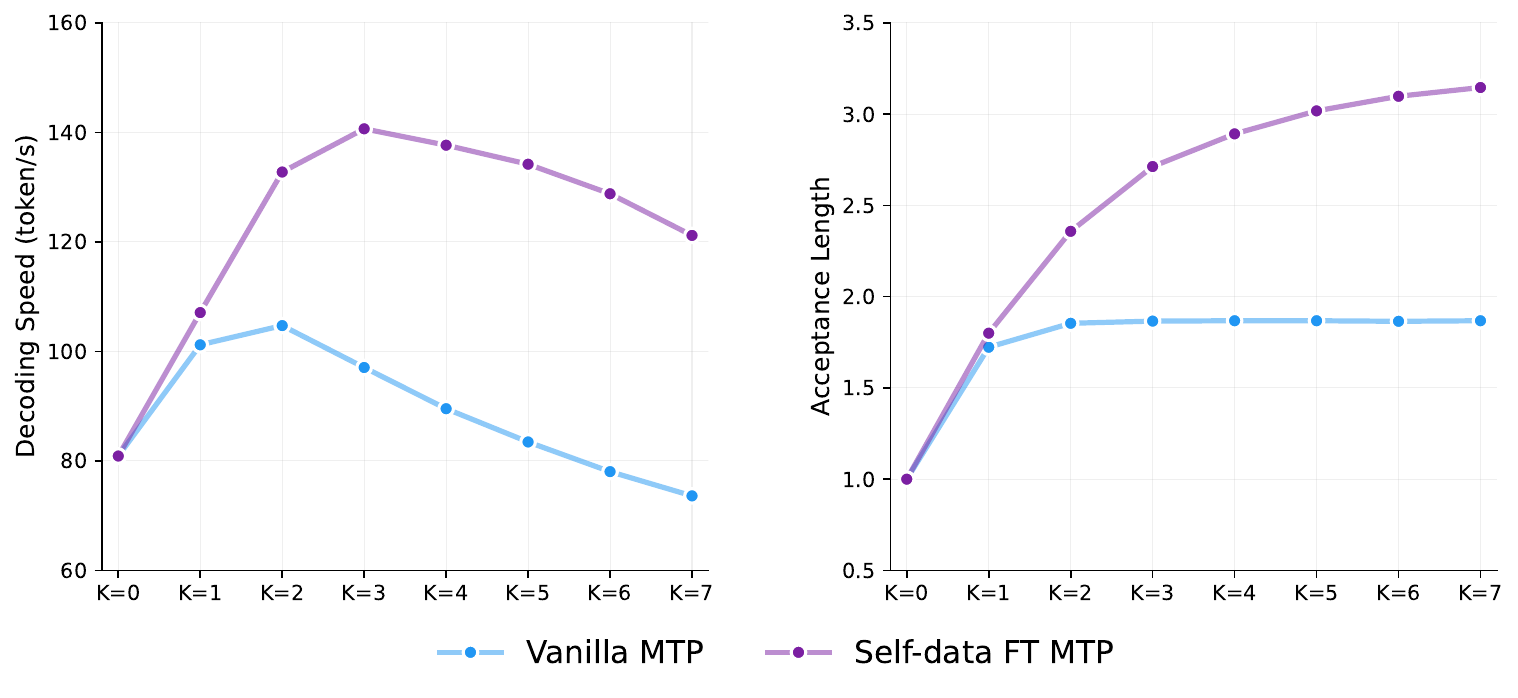}
    \caption{Decoding Speed and Acceptance Length with different MTP draft token counts on a single A100.}
    \label{fig:num_mtp}
\end{figure}

From Figure \ref{fig:num_mtp}, we can see that FastMTP achieves peak speedup at $K=3$, reaching 140 token/s, while the vanilla MTP checkpoint peaks earlier at $K=2$ with lower output throughput. For the acceptance length, FastMTP with self-distilled fine-tuning shows consistently growing acceptance length from 1.0 to 3.2 as $K$ increases, demonstrating that our training strategy successfully enables the model to learn multi-step predictions. In contrast, vanilla MTP maintains a nearly flat acceptance length around 1.8 from $K=2$, indicating poor prediction quality beyond the first position.

Despite the monotonically increasing acceptance length, the decoding speed peaks at $K=3$ and then gradually declines. While longer draft sequences increase the average acceptance length, they also incur additional computational overhead. Beyond $K=3$, the marginal gains in acceptance length cannot compensate for the growing computational overhead of generating longer drafts, as distant tokens become progressively harder to predict accurately, which reduces overall efficiency. Therefore, \textbf{FastMTP achieves maximum speedup at $\mathbf{K=3}$ draft length, where the acceptance length improvements through training optimally outweigh the computational overhead of recursive drafting.}

% \begin{figure}[htbp]
%     \centering
%     \includegraphics[width=0.65\linewidth]{./figures/loss.pdf}
%     \caption{Training Loss of MTP at different draft steps.}
%     \label{fig:loss}
% \end{figure}

\subsubsection{Vocabulary compression trade-offs}

To investigate the impact of vocabulary compression on performance, we evaluated four vocabulary sizes: $|\mathcal{V}_{high}|=\{8\text{k}, 16\text{k}, 32\text{k}, 64\text{k}\}$, compared to the full 152k-token vocabulary. We selected two representative benchmarks: C-Eval~\citep{huang2023ceval} for Chinese knowledge question answering and MT-Bench~\citep{zheng2023judging} for English multi-turn conversation. This selection helps us to analyze language-specific compression effects across diverse linguistic contexts.

Vocabulary frequency patterns are observed to be highly language-dependent, confirming FR-Spec's context-related acceleration paradigm~\citep{zhao2025frspec}. The original implementation uses token statistics from SlimPajama-627B~\citep{daria2023slimpajama}, an English-dominated corpus, which yields low acceptance rates and consequently poor performance on real-world Chinese tasks due to mismatched frequency distributions. To address this limitation, we construct language-specific compressed vocabularies tailored to each target domain. For Chinese tasks, we select the Chinese-DeepSeek-R1-Distill-110k-SFT dataset~\citep{liu2025chinese}, which contains high-quality mathematical reasoning and diverse Chinese instruction-following examples distilled from the powerful DeepSeek-R1 model~\citep{deepseek-ai2025deepseekr1}, to identify high-frequency tokens characteristic of Chinese generation patterns. During inference, FastMTP dynamically switches to appropriate compressed vocabularies according to the input context.

Table \ref{tab:fr} presents the average acceptance length, decoding speed, and speedup ratios across different vocabulary configurations. The results demonstrate that even aggressive vocabulary compression incurs only minimal degradation in acceptance length while delivering considerable speedup gains. Notably, the optimal compression ratio varies across languages. For MT-Bench, the 32k vocabulary configuration achieves peak performance with 2.028$\times$ speedup, sacrificing merely 0.068 tokens in acceptance length compared to the full vocabulary. This aligns with findings from prior work on English-centric compression. In contrast, C-Eval achieves optimal performance with a more aggressive 16k vocabulary, yielding 1.990$\times$ speedup with an even smaller acceptance length reduction of 0.026 tokens. This suggests that Chinese text generation may exhibit more concentrated token usage patterns, making it benefit more substantially from vocabulary compression strategies. In conclusion, \textbf{the optimal vocabulary size differs by language—16k for Chinese, 32k for English—revealing distinct token distribution patterns.}

\begin{table}[htbp]
\centering
\caption{Different FR-Spec configurations.}
\label{tab:fr}
\begin{tabular}{clccc}
\toprule
\multicolumn{1}{l}{}       &                                          & \textbf{$\tau$}                             & \textbf{token/s}                        & \textbf{speedup}                       \\ \midrule
                           & Baseline                                 & 1.000                                  & 31.627                                  & 1.000x                                  \\
                           & Full Vocab. (152k)                    & 2.551                                  & 54.364                                  & 1.719x                                  \\
                           & +FR 8k                                   & 2.448                                  & 62.015                                  & 1.961x                                  \\
                           & \cellcolor[HTML]{E5F6FF}\textbf{+FR 16k} & \cellcolor[HTML]{E5F6FF}\textbf{2.525} & \cellcolor[HTML]{E5F6FF}\textbf{62.928} & \cellcolor[HTML]{E5F6FF}\textbf{1.990x} \\
                           & +FR 32k                                  & 2.549                                  & 62.210                                  & 1.967x                                  \\ 
\multirow{-6}{*}{C-Eval}   & +FR 64k                                  & 2.550                                  & 60.083                                  & 1.900x                                  \\ \midrule
                           & Baseline                                 & 1.000                                  & 31.276                                  & 1.000x                                  \\
                           & Full Vocab. (152k)                    & 2.690                                  & 56.333                                  & 1.801x                                  \\
                           & +FR 8k                                   & 2.426                                  & 60.260                                  & 1.927x                                  \\
                           & +FR 16k                                  & 2.519                                  & 62.144                                  & 1.987x                                  \\
                           & \cellcolor[HTML]{E5F6FF}\textbf{+FR 32k} & \cellcolor[HTML]{E5F6FF}\textbf{2.622} & \cellcolor[HTML]{E5F6FF}\textbf{63.420} & \cellcolor[HTML]{E5F6FF}\textbf{2.028x} \\
\multirow{-6}{*}{MT-Bench} & +FR 64k                                  & 2.677                                  & 61.572                                  & 1.969x     \\ \bottomrule                             
\end{tabular}
\end{table}

\subsection{Ablation experiments}

We conduct comprehensive ablation experiments to validate the contribution of each component in FastMTP and analyze the acceptance rates of each draft position after training.

\paragraph{Impact of Design Choices.} To quantify the individual contributions of our design choices, we evaluate three configurations against the vanilla MTP baseline: fixed-data fine-tuning, self-distilled data fine-tuning, and the complete FastMTP with vocabulary compression. As shown in Table~\ref{tab:all}, FastMTP achieves a 2.03$\times$ speedup, representing an 82\% improvement over vanilla MTP (1.21$\times$), a 36\% gain over fixed-data fine-tuning (1.67$\times$), and a 22\% increase over self-distilled data fine-tuning without vocabulary compression (1.81$\times$). These incremental improvements validate the effectiveness of each component: fine-tuning the shared-weight MTP head enables more accurate and faster multi-token drafting during inference, self-distillation achieves better distribution alignment between draft and main model for higher acceptance length, and vocabulary compression reduces computational overhead to increase output throughput with minimal impact on acceptance rates.

\begin{figure}[htbp]
    \centering
    \begin{subfigure}[b]{0.58\textwidth}
        \centering
        \includegraphics[width=\textwidth]{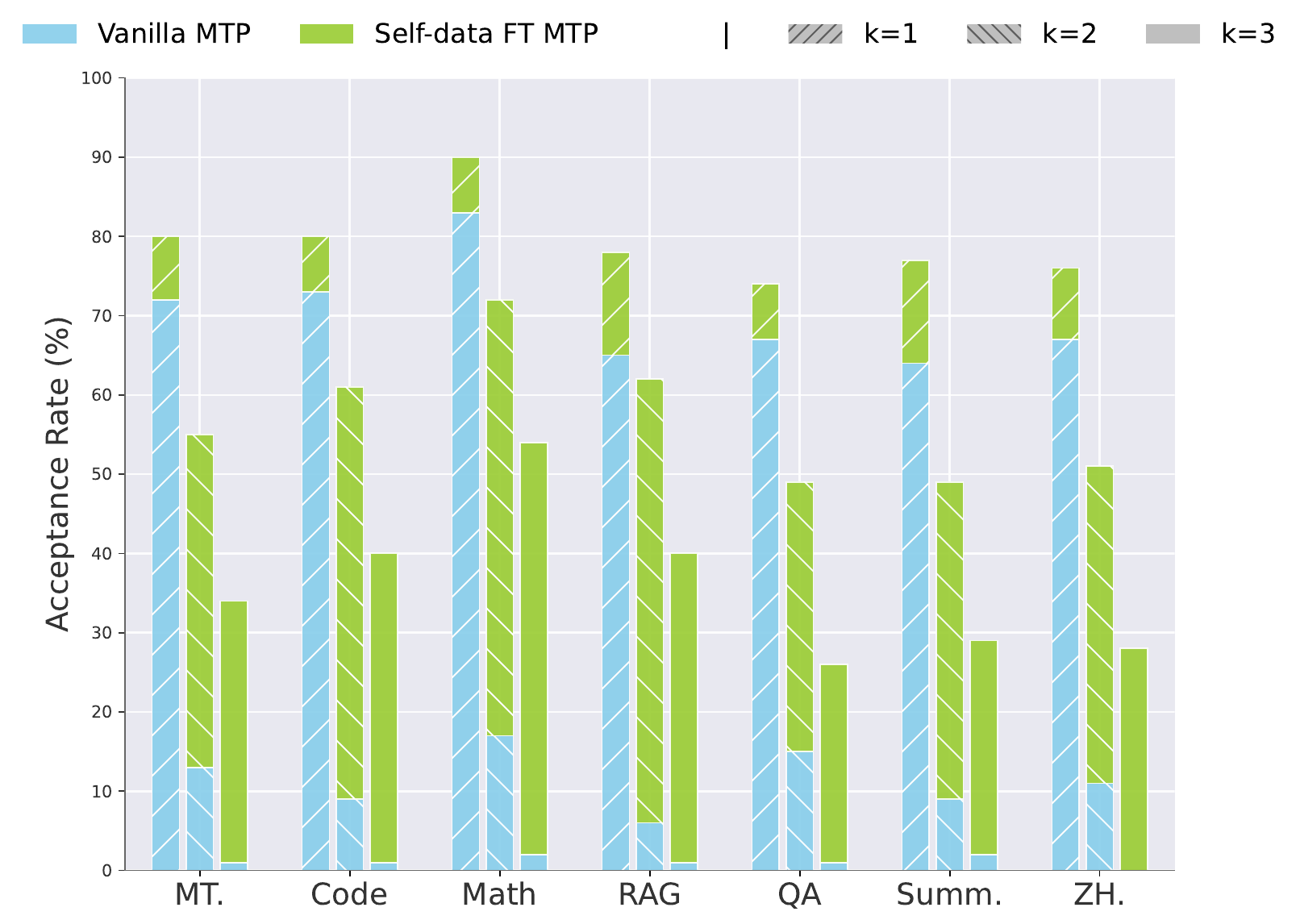}
        \caption{}
        \label{fig:acc_rate}
    \end{subfigure}
    \hfill
    \begin{subfigure}[b]{0.4\textwidth}
        \centering
        \includegraphics[width=\textwidth]{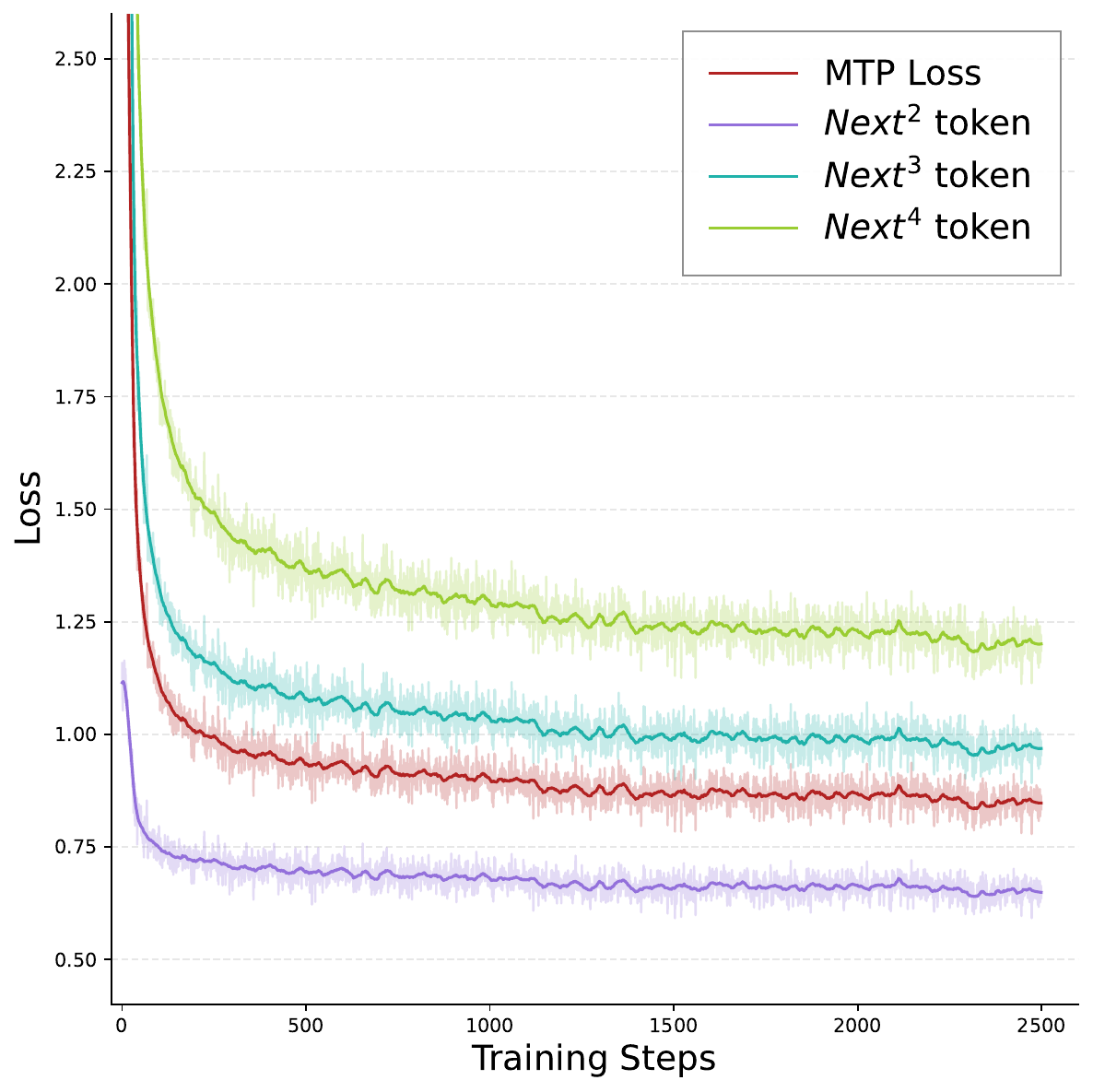}
        \caption{}
        \label{fig:loss}
    \end{subfigure}
    \caption{(a) Acceptance rate improvement of self-distilled dataset fine-tuned MTP compared with vanilla MTP at different draft steps. (b) Training Loss of MTP at different draft steps.}
    \label{fig:acc_rate_loss}
\end{figure}

\paragraph{Acceptance Rate Analysis.} Figure \ref{fig:acc_rate} further demonstrates the critical role of fine-tuning in enabling effective multi-token drafting. Vanilla MTP exhibits severe performance degradation beyond the first draft step, with acceptance rates dropping sharply from approximately 70\% at $k=1$ to merely 10\% at $k=2$, and approaching zero at $k=3$. This collapse exposes the inherent limitation of deploying vanilla MTP checkpoints for recursive multi-token prediction in speculative decoding—it was trained only for single-step ahead prediction and thus struggles to maintain high acceptance rates when applied to deeper draft positions. In contrast, our fine-tuned MTP achieves substantially higher acceptance rates across all draft steps: 80\% at $k=1$, 56\% at $k=2$, and 36\% at $k=3$ on average. Mathematical reasoning shows the highest gains, maintaining over 50\% acceptance rate even at $k=3$ compared to vanilla MTP's mere 3\%. The consistent superiority across all seven evaluated tasks validates that fine-tuning with self-distilled data effectively equips the MTP head with the ability to capture dependencies among consecutive future tokens essential for effective multi-step prediction. Figure~\ref{fig:loss} illustrates the training loss curves. The loss drops sharply in the early training stages (approximately 0.5 epochs), followed by slower but steady improvement until convergence. As expected, predictions become more inaccurate as the token position increases, with deeper positions inevitably showing higher losses, further illustrate that distant tokens become progressively harder to predict accurately.

% To further illustrate that distant tokens become progressively harder to predict accurately, we present the training loss curves for each step when training with the maximum prediction depth $K=3$ in Figure \ref{fig:loss}. The loss drops sharply in the early training stages (approximately 0.5 epochs), followed by slower but steady improvement until convergence. As expected, predictions become more inaccurate as the token position increases, with deeper positions inevitably showing higher losses.

\section{Related Work}

\subsection{LLM inference acceleration} 

The growing computational demands of LLMs have motivated extensive research into inference acceleration techniques, which can be broadly categorized into four main directions: architectural innovations, model compression, framework optimizations, and speculative decoding. 

Architectural innovations focus on redesigning model components for enhanced efficiency. Representative works include efficient attention mechanisms that reduce quadratic complexity, such as linear attention~\citep{katharopoulos2020transformers,yang2025parallelizing,yang2024gated} that approximates softmax attention as linear operations, sparse attention~\citep{child2019generating,lu2025moba} that restricts computation to selective key subsets based on fixed or dynamic sparsity patterns~\citep{sun2025efficient}, and low-rank attention that employs linear projection for dimensional compression~\citep{deepseek-ai2024deepseekv2}. Model compression techniques reduce computational and memory requirements, including quantization~\citep{frantar2023gptq,lin2024awq,dettmers2022llmint8,xiao2024smoothquant}, pruning~\citep{frantar2023sparsegpt,sun2024simple,ma2023llmpruner}, and knowledge distillation~\citep{gu2024minillm,hsieh2023distilling,ho2023large}. Framework optimizations improve deployment efficiency through system-level engineering. Notable contributions include FlashAttention~\citep{dao2022flashattention,dao2023flashattention2,shah2024flashattention3} that fuses the entire attention operation into a single operator, vLLM with PagedAttention~\citep{kwon2023efficient} for memory management, SGLang with RadixAttention~\citep{zheng2024sglang} for KV cache reuse, and TensorRT-LLM~\citep{nvidia2023tensorrtllm} with built-in support for various parallelism strategies and advanced features. Speculative decoding~\citep{leviathan2023fast,chen2023accelerating,stern2018blockwise} leverages draft-then verification paradigms to accelerate decoding.

\subsection{Speculative decoding optimization}

Among the aforementioned acceleration techniques, speculative decoding uniquely preserves output quality while achieving significant speedups. Optimization efforts in speculative decoding primarily target two aspects: obtaining higher acceptance rates, and reducing draft generation overhead. 

To achieve higher acceptance rates, prior research has explored two primary directions. The first focuses on improving draft quality through specialized model design and training. Several works develop self-drafting mechanism by adding additional prediction heads to the target model. Medusa~\citep{cai2024medusa} employs extra MLP heads that reuse the last hidden states from the target model to predict the next few tokens in parallel. In contrast, EAGLE~\citep{li2024eagle,li2024eagle2,li2025eagle3} incorporates both the last hidden state and preceding tokens to draft in an autoregressive way, significantly improving draft stability and accuracy, establishing it as the current state-of-the-art approach. Other works utilize smaller models from the same model series as draft model~\citep{leviathan2023fast,chen2025cascade}, exploiting similarity for better alignment. The second direction slightly relaxes the matching requirement to trust the drafting results more, leading to higher acceptance of drafted tokens~\citep{xia2024unlocking}. For instance, SpecDec~\citep{xia2023speculative} only requires the drafted tokens to fall in top-$\beta$ candidates with a tolerable score gap away from the top-1 result.

To reduce draft generation overhead, recent works have explored various optimization strategies. One line of work focuses on dynamic draft tree construction. BiLD~\citep{kim2023speculative} and Kangaroo~\citep{liu2024kangaroo} implement early stopping mechanisms based on draft model confidence to control tree depth, while EAGLE-2~\citep{li2024eagle2} goes further by leveraging confidence scores to approximate acceptance rates and adjust the draft tree structures accordingly. Another line of work has targeted computational bottlenecks in the drafting process itself. TriForce~\citep{sun2024triforce} accelerates long-context drafting through KV-cache compression, while Ouroboros~\citep{zhao2024ouroboros} enhances efficiency by adapting lookahead decoding~\citep{fu2024break} techniques. Addressing vocabulary-related overhead, FR-Spec~\citep{zhao2025frspec} identifies the computational bottleneck of large-vocabulary LM heads and restricts the drafting space to high-frequency token subsets to make draft models faster.

% Recent advances have also integrated Mixture-of-Experts (MoE)~\citep{shazeer2017outrageously} technique into LLMs, where routing modules dynamically activate specialized experts per token, achieving efficiency through sparse computation while maintaining model capacity.
% One line of work focuses on reducing CoT redundancy, including reinforcement learning with length penalties~\citep{luo2025o1pruner,aggarwal2025l1} and supervised fine-tuning on variable-length CoT data~\citep{ma2025cotvalve}. Another line of work explores techniques such as efficient attention mechanisms~\citep{katharopoulos2020transformers,child2019generating,yang2024gated,lu2025moba} and model compression~\citep{lin2024awq,xiao2024smoothquant,gu2024minillm,hsieh2023distilling} to reduce computational overhead.

\section{Conclusion}

In this work, we propose FastMTP as an improvement over vanilla multi-token prediction in the speculative decoding during large language models inference. FastMTP introduces two key implementations. First, we train a single shared-weight MTP head through self-distillation, eliminating the need for multiple independent MTP modules while improving multi-step prediction capability. Second, we employ language-aware vocabulary compression to reduce computational costs during draft generation. Experimental results demonstrate the superiority of the proposed method, where both draft token quality and overall output speedup can be enhanced in a series of scenarios.

\bibliographystyle{colm2024_conference}
\bibliography{main}

\newpage

\appendix

\section{Training Datasets Details} \label{app:train_data}

We collect our dataset for MTP head fine-tuning from a variety of data sources. To ensure distribution alignment, we employ self-distillation where all responses are generated by the main model itself. For each prompt $x_n$ extracted from the source datasets, we generate corresponding response $\tilde{y}_n$ using the main model with the following generation configurations: temperature of 0.6, top-k of 20, top-p of 0.95, and maximum length of 4096 tokens. The entire data distillation process is conducted using the SGLang inference framework~\citep{zheng2024sglang}. 

After distillation, we apply de-duplication, data cleaning, and mixing strategies to curate high-quality tokens:
\begin{itemize}
    \item \textbf{De-duplication:} We perform global MinHash de-duplication both within individual data sources and across the entire dataset to remove near-duplicate samples.
    
    \item \textbf{Data cleaning:} We develop heuristics to filter out low-quality samples. Some examples of heuristics include: (1) samples with incomplete or truncated reasoning chains; (2) samples containing excessive repetitive content; and (3) samples that fall outside desired length ranges.
    
    \item \textbf{Data mixing:} Our final dataset comprises four major categories with carefully balanced proportions: approximately 42\% of tokens corresponding to general knowledge and tasks, 18\% to mathematical and reasoning content, 13\% to code, and 27\% to Chinese texts, yielding the final 389.4K high-quality training examples.
\end{itemize}

\section{Evaluation Benchmarks Details} \label{app:bench}

We evaluate FastMTP across seven diverse tasks to ensure comprehensive coverage of real-world applications: multi-turn conversation, code generation, mathematical reasoning, retrieval-augmented generation (RAG), question answering, summarization, and Chinese knowledge assessment. Adapting from the Spec-Bench evaluation framework~\citep{xia2024unlocking}, we carefully select benchmarks that represent distinct generation patterns and challenges. To ensure fair evaluation, we randomly sample 80 instances from each task (102 for C-Eval) and avoid benchmarks potentially exposed during MTP training.
% Table \ref{tab:eval} summarizes the evaluation setup.

Our benchmark suite comprises:
\begin{itemize}
    \item \textbf{MT-Bench}~\citep{zheng2023judging}: A series of open-ended questions that evaluate a chatbot’s multi-turn conversational and instruction-following ability. MT-bench is also carefully constructed to differentiate chatbots based on their core capabilities, such as reasoning and math.
    
    \item \textbf{LiveCodeBench-v6}~\citep{jain2024livecodebench}: A comprehensive and contamination-free evaluation of LLMs for code collects new problems over time from contests across three competition platforms, namely LeetCode, AtCoder, and CodeForces. We select the sixth version of the dataset, which contains 1055 problems released between May 2023 and Apr 2025.
    
    \item \textbf{MATH-500}~\citep{lightman2023lets}: A dataset comprising 500 challenging mathematical problems designed to test advanced mathematical reasoning and problem-solving skills. It includes problems from various domains such as algebra, calculus, geometry, and number theory, primarily at high school and early undergraduate levels. 
    % We select this dataset to avoid potential contamination from GSM8K which may have been exposed during MTP head training.
    
    \item \textbf{Natural Questions}~\citep{kwiatkowski2019natural}: A large-scale QA dataset containing real user queries paired with high-quality annotations from Wikipedia documents. We utilize this benchmark in two subtasks: question answering and retrieval-augmented generation (RAG).
    
    \item \textbf{CNN/Daily Mail}~\citep{nallapati2016abstractive}: An English-language dataset containing just over 300k unique news articles as written by journalists at CNN and the Daily Mail, each with 3-4 highlights that summarize the contents of the article. This benchmark evaluates the model's ability to produce extractive and abstractive summarizations of news articles while preserving key information.
    
    \item \textbf{C-Eval}~\citep{huang2023ceval}: A comprehensive Chinese evaluation suite designed to assess advanced knowledge and reasoning abilities of foundation models in a Chinese context. C-Eval comprises multiple-choice questions across four difficulty levels: middle school, high school, college, and professional. The questions span 52 diverse disciplines, ranging from humanities to science and engineering. 
\end{itemize}

% \begin{table}[htbp]
% \centering
% \caption{Sources for MTP evaluation dataset}
% \label{tab:eval}
% \begin{tabular}{lllc}
% \toprule
% \textbf{Subtask}          & \textbf{Dataset}  & \textbf{Source} & \multicolumn{1}{l}{\textbf{Samples Used}} \\ \midrule
% Multi-turn Conversation   & MT-bench          & \citep{zheng2023judging}                & 80                                        \\
% Retrieval-aug. Generation & Natural Questions & \citep{kwiatkowski2019natural}                & 80                                        \\
% Summarization             & CNN/Daily Mail    & \citep{nallapati2016abstractive}                & 80                                        \\
% Question Answering        & Natural Questions & \citep{kwiatkowski2019natural}                & 80                                        \\
% Mathematical Reasoning    & MATH-500          & \citep{lightman2023lets}                & 80                                        \\
% Coding                    & LiveCodeBench-v6  & \citep{jain2024livecodebench}                & 80                                        \\
% Chinese Knowledge         & C-Eval            & \citep{huang2023ceval}                & 102                                       \\ \midrule
% \textbf{Total}            &                   &                 & 582      \\ \bottomrule                                
% \end{tabular}
% \end{table}

\end{document}

%% file: math_commands.tex
\usepackage{amsmath,amsfonts,bm}

\def\eqref#1{equation~\ref{#1}}
\def\1{\bm{1}}

\DeclareMathAlphabet{\mathsfit}{\encodingdefault}{\sfdefault}{m}{sl}
\SetMathAlphabet{\mathsfit}{bold}{\encodingdefault}{\sfdefault}{bx}{n}